\documentclass{article}
\usepackage{spconf,amsmath,graphicx}
\usepackage{subfigure}

\title{Liver lesion segmentation informed by joint liver segmentation}

\name{Eugene Vorontsov\textsuperscript{1,2}, An Tang\textsuperscript{3}, Chris Pal\textsuperscript{1,2}, Samuel Kadoury\textsuperscript{1,3} \thanks{We thank Nvidia for GPU donation to our lab.}}
\address{\textsuperscript{1} \'Ecole Polytechnique de Montr\'eal\\
\textsuperscript{2} Montreal Institute for Learning Algorithms (MILA)\\
\textsuperscript{3} Centre de recherche du CHUM (CRCHUM)}

\begin{document}
\maketitle

\begin{abstract}
We propose a model for the joint segmentation of the liver and liver lesions in computed tomography (CT) volumes. We build the model from two fully convolutional networks, connected in tandem and trained together end-to-end. We evaluate our approach on the 2017 MICCAI Liver Tumour Segmentation Challenge, attaining competitive liver and liver lesion detection and segmentation scores across a wide range of metrics. Unlike other top performing methods, our model output post-processing is trivial, we do not use data external to the challenge, and we propose a simple single-stage model that is trained end-to-end. However, our method nearly matches the top lesion segmentation performance and achieves the second highest precision for lesion detection while maintaining high recall.
\end{abstract}


\section{Introduction}
\label{sec:intro}

The segmentation of liver tumours tumours in computed tomography (CT) is required for assessment of tumour load, treatment planning, prognosis, and monitoring of treatment response. Because manual segmentation is time consuming, tumour size is usually estimated in clinical practice from measurements in the axial plane of the largest diameter of the tumour and the diameter perpendicular to it \cite{eisenhauer2009new}. Nevertheless, tumour volume is a better predictor of patient survival than diameter \cite{chapiro2014identifying}. Hence, there is a clear need for tools to aid with tumour detection and segmentation.

Recent advances in computer vision have spurred the resurgence and refinement of deep neural networks which can now exceed human performance in object classification from natural images \cite{ioffe2015batch}. Exploration of this promising avenue has only recently begun for medical image segmentation. Current models \cite{drozdzal2016importance, christ2016automatic, milletari2016v, poudel2016recurrent, chen2016deep} are based on fully convolutional neural networks (FCN) \cite{long2015fully, hariharan2015hypercolumns}, often similar to the UNet \cite{ronneberger2015u}. We exploit the architecture that is evaluated in \cite{drozdzal2016importance} to construct a model configuration for segmenting metastatic lesions in the liver within CT volumes.

We attain competitive liver and liver lesion detection and segmentation scores across a wide range of metrics in the 2017 MICCAI Liver Tumour Segmentation Challenge (LiTS). Unlike other top scoring methods, we do not pre-process the data, we employ only trivial post-processing of model outputs, and we propose a single-stage model that is trained end-to-end.

\section{Method}
\label{sec:method}

\begin{figure*}[htb!]
\centering
\subfigure{
\includegraphics[width=0.95\textwidth]{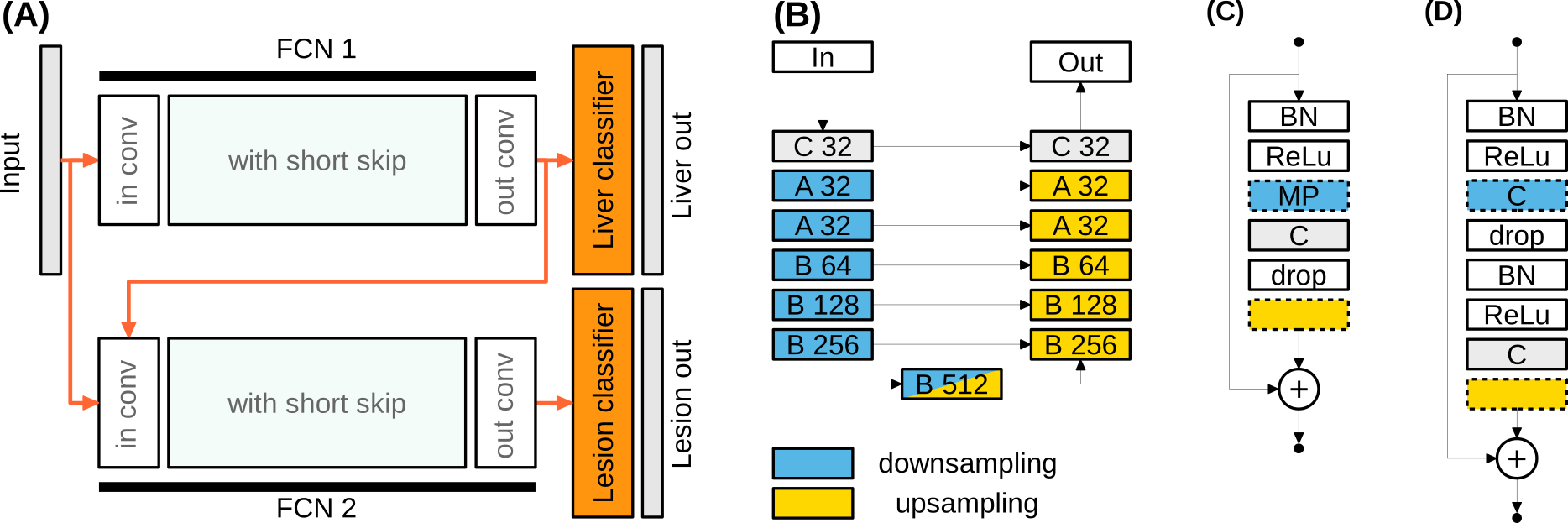}
}
\caption{(A) Two FCNs, FCN 1 and 2, each take a 2D axial slice as input. FCN 1 produces a segmentation mask for the liver; FCN 2 for lesions. The latent representation produced by FCN 1 is passed as an additional input to FCN 2. (B) FCN structure with the number of convolution filters noted in each block. Blocks coloured blue perform downsampling while those coloured yellow perform upsampling. ``C" denotes a 3x3 pixel convolution layer; ``A" and ``B" denote blocks A and B, shown in (C) and (D), respectively. ``BN", ``ReLU", and ``MP", denote batch normalization, rectified linear units, and max pooling, respectively. Blocks with dashed lines are used in only the upsampling or the downsampling path, as denoted by colour.}
\label{fig:network}
\end{figure*}

\subsection{Model}

We construct a model with two fully convolutional networks (FCNs), one on top of the other, trained end-to-end to segment 2D axial slices. Both networks are UNet-like \cite{ronneberger2015u} with short and long skip connections as in \cite{drozdzal2016importance}. The combined network is shown in Figure \ref{fig:network} (A). FCN 1 takes an axial slice as input and its output is passed to a linear classifier that outputs (via a sigmoid) a probability for each pixel being within the liver. FCN 2 takes as input both the axial slice and the output of FCN 1. The input thus has a number of channels equal to the number of channels in the representation produced by FCN 1 plus one channel which contains the axial slice. The representation produced by FCN 1 is effectively passed to every layer of FCN 2 due to short skip connections, after first passing through the first convolution layer of FCN 2. The output representation of FCN 2 is passed to a lesion classifier, of the same type as the liver classifier.
  
The FCN 1 and 2 networks have an identical architecture, as shown in Figure \ref{fig:network} (B).  In each FCN, an input passes through an initial convolution layer and is then processed by a sequence of convolution blocks at decreasing resolutions and an increasing receptive field size. This contracting path is shown in blue on the left. An expanding path (right, in yellow) then reverses the downsampling performed by the contracting path. The expanding path mirrors the structure of the contracting path. Each block in the expanding path takes as input the sum of the previous block's output and the output of its corresponding block from the contracting path; this allows the expanding path to recover spatial detail lost with downsampling. Representations are thus skipped from left to right along long skip connections. 
  
We used two types of blocks: block A and block B. Both have short skip connections which sum the block's input into its output, as shown in Figure \ref{fig:network} (C), (D). Both blocks contain dropout layers, a downsampling layer when used along the contracting path, and an upsampling layer when used along the expanding path. The downsampling layer in block A is max pooling. In block B it is basic grid subsampling, achieved by applying convolutions with a stride of 2. The upsampling layer performs simple nearest neighbour interpolation. The main difference between blocks A and B is in the number of convolution operations: block A contains one convolution layer and block B contains 2. All convolution layers use 3x3 filters; the number of filters is shown for each block in Figure \ref{fig:network} (B).

\subsection{Data Set}
\label{sec:data}

The proposed segmentation method was applied to metastatic lesions in the liver imaged with CT. The dataset included 200 CT volumes with variable coverage, either limited to the abdomen or including the entire abdomen and thorax. All volumes were enhanced with a contrast agent, imaged in the portal venous phase. All volumes contained a variable number of axial slices with a resolution of 512x512 pixels, with varying slice thicknesses. Of the 200 volumes, 130 volumes were provided publicly with manual segmentations of the liver and liver lesions while 70 were withheld until near the end of the LiTS challenge for evaluation. Manual segmentations were not provided for this evaluation set.

Of the 130 cases with segmentations, we used 115 for training and 15 for validating our segmentation models. We did not apply any pre-processing to the images except for basic image-independent scaling of the intensities to ensure inputs to our neural networks were within a reasonable range: we divided all pixel values by 255 and then clipped the resulting intensities to within [-2, 2].

\subsection{Training the Model}
\label{sec:training}

We trained the model only on 2D axial slices that contain the liver, using RMSprop \cite{Tieleman2012} and the Dice loss defined in \cite{drozdzal2016importance,milletari2016v}. For data augmentation, we applied random horizontal and vertical flips, rotations up to 15 degrees, zooming in and out up to 10\%, and elastic deformations as described by \cite{ronneberger2015u}. In order to improve training time, allowing us to test many models and hyperparameters in a short time, we first downscaled all slices from a 512x512 resolution to 256x256. This initial model was trained with a 0.001 learning rate (0.9 momentum). The model was then fine-tuned on full resolution slices, using a 0.0001 learning rate.

\begin{table*}[htb!]
\centering
\begin{tabular}{ ccccccc } 
  \multicolumn{7}{l}{\textbf{Segmentation}} \\
  & Dice & VOE (\%) & RVD (\%) & ASSD (mm) & MSD (mm) & RMSD (mm) \\
  leHealth & - & 39.4 & 5.921 & 1.189 & 6.682 & 1.726 \\
  hchen & - & 35.6 & 5.164 & 1.073 & 6.055 & 1.562 \\
  hans.meine & - & 38.3 & 0.464 & 1.143 & 7.322 & 1.728 \\
  \textbf{our} & 0.773 & 35.7 & 12.124 & 1.075 & 6.317 & 1.596 \\
  
  \hline
  
  \textbf{Detection} & \multicolumn{2}{l}{\textbf{$>$50\% overlap}} &
  \multicolumn{2}{l}{\textbf{$>$0\% overlap}} &
  \multicolumn{2}{l}{\textbf{Mixed measures}} \\
  & Precision & Recall & Precision & Recall & Global Dice & Dice per case \\
  leHealth & 0.156 & 0.437 & - & - & 0.794 & 0.702 \\
  hchen & 0.409 & 0.408 & - & - & 0.8290 & 0.686 \\
  hans.meine & 0.496 & 0.397 & - & - & 0.796 & 0.676 \\
  \textbf{our} & 0.446 & 0.374 & 0.686 & 0.574 & 0.783 & 0.661 \\

\end{tabular}
\caption{Segmentation and detection metrics evaluated for the proposed method on the MICCAI LiTS 2017 test set.}
\label{tab:results}
\end{table*}

The model was trained for 200 epochs on downscaled slices (batch size 40) and fine-tuned for 30 epochs on full-resolution slices (batch size 10). The final model weights were those which yielded the best loss on the validation set.

The proposed model is limited to processing 2D slices due to memory constraints. To improve segmentation performance and consistency across slices for the LiTS challenge, we introduced some cross-slice context. For every slice, three consecutive slices were considered (one above, one below). The pre-classifier outputs from each of the three slices were combined by a convolution (3x3 kernel); a new classifier for the middle slice was trained on the resultant features.

\subsection{Generating Segmentations}
\label{sec:testing}

At test time, segmentation predictions were averaged across all four input orientations achieved by vertical and horizontal flips. This was done for three similar models and the predictions of the ensemble were averaged. A liver segmentation was extracted by selecting the largest connected component in the model's liver segmentation prediction. A lesion segmentation was extracted by cropping the model's lesion segmentation prediction to a dilated version of the liver segmentation. For dilation, we chose to extend the liver's boundaries by 20mm. This eliminated false positives outside of the liver without incorrectly cropping out lesions when the liver is slightly under-segmented. Beyond cropping to a single liver, no post-processing was performed on the model outputs.

\section{Results and Discussion}
\label{sec:results}

\begin{figure}[h!]
\centering
\subfigure{
\includegraphics[width=0.45\textwidth]{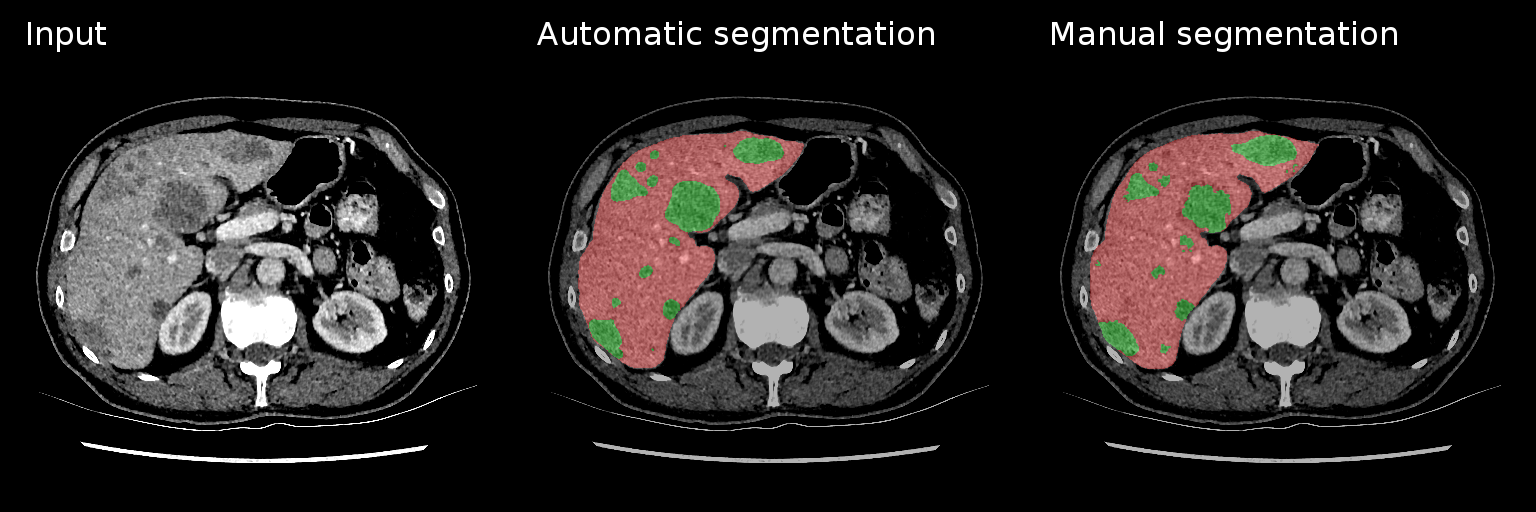}
}
\caption{Example of segmentation output compared to ground truth ("Manual segmentation"). Lesions in green, liver in red.}
\label{fig:example}
\end{figure}

The proposed method performed relatively well in the MICCAI LiTS challenge, achieving similar scores to other top methods, as shown in Table \ref{tab:results} (example segmentation in Figure \ref{fig:example}. Segmentation metrics evaluate the segmentation of detected lesions (averaged across lesions). They are comprised of a per-lesion Dice score, a volume overlap error (VOE), a relative volume difference (RVD), the average symmetric surface distance (ASSD), the maximum surface distance (MSD), and the root means square symmetric surface distance (RMSD). Detection metrics were evaluated as precision and recall at $>$50\% and $>$0\% overlap (measured by intersection over union) of each predicted lesion with the corresponding ground truth. Dice metrics that confound both detection and segmentation were the Dice score computed on all combined volumes (global Dice) and the mean Dice score per volume (Dice per case). Entries in the challenge were ranked according to the Dice per case, placing our method fourth in lesion segmentation with a score of 0.661. Liver segmentation performed well, with an average Dice per case of 0.951 (the best entry scored 0.963).

Although three methods attained higher mean Dice per case, our method compares favourably in terms of higher detection scores or lower complexity. While \textit{leHealth} attained the top Dice per case score of 0.702, the method suffers from very low precision (0.156 compared to our 0.446, at $>$50\% overlap). It also relies on extensive model ensembling and post-processing. The second method, labeled \textit{hchen} in Table \ref{tab:results}, attained a Dice per case of 0.686 but at the cost of a lower precision (0.409 at $>$50\% overlap) \cite{li2017h}. This method relies on a three-stage process where the liver is first roughly segmented, then liver and lesions are segmented with a 2D FCN, and finally, the segmentations are refined with a small 3D FCN that takes the initial segmentation predictions as input. The authors found that using a pre-trained 2D model significantly boosted performance. By contrast, we developed a single-stage pipeline in which we did not use pre-trained models; we will extend our method to 3D in the future. Finally, \textit{hans.meine} (using the approach described in \cite{chlebus2017neural}) attained a Dice per case of 0.676 with detection scores at 50\% overlap that are slightly higher than for our method; however, that method involved post-processing with a random forest classifier to improve precision and used data other than that provided in the challenge to train a liver segmentation model. In comparison, our post-processing was trivial and we trained our model on only the data provided in the challenge.

All top methods used an FCN for lesion segmentation, conditioned on prior liver segmentation. In this regard, our approach differs only in that it is a single-stage model, trained end-to-end, and the lesion segmentation (FCN 2) is conditioned on the high-dimensional pre-classifier representation in the liver (FCN 1), rather than on the liver classifier outputs. This configuration allows FCN 2 to focus on the liver when performing lesion segmentation and ignore lesions far from the liver. We found that using a single FCN to segment lesions and the liver simultaneously is less effective. This may be because this does not model the dependence of the lesion segmentations on that of the liver. In addition, training FCN 1 and 2 end-to-end allows FCN 1 to learn a representation that is amenable for lesion segmentation, boosting the performance of FCN 2. Indeed, \cite{drozdzal2017learning} found that an FCN may act as an effective learned pre-processor for another FCN.

\section{Conclusion}

The proposed model performs end-to-end joint liver and lesion segmentation in CT quickly without any need for pre-processing of input images or complicated post-processing of the outputs. Segmentation performance could be improved by extending the proposed model to processing the whole CT volume rather than slice inputs. The proposed model's simplicity makes it a good base model for architectural research toward improving liver and liver lesion segmentation.

\section{Acknowledgements}

We thank An Tang and Gabriel Chartrand for preparing some of the data used in the LiTS challenge.

\bibliographystyle{IEEEbib}
\bibliography{strings,refs}

\end{document}